\documentclass[letterpaper, 10 pt, conference]{ieeeconf}  

\IEEEoverridecommandlockouts                              
\usepackage{graphicx}      
\usepackage{amsmath}
\usepackage{amssymb}
\usepackage{stackengine}
\usepackage{url}
\usepackage{cite}

\DeclareMathOperator*{\argmin}{arg\;min}

\title{\LARGE \bf
Improved Initialization of State-Space Artificial Neural Networks
}

\author{Maarten Schoukens
\thanks{This research was supported by the European Union's Horizon 2020 research and innovation programme under the Marie Sklodowska-Curie Fellowship (grant agreement nr. 798627).}
\thanks{M. Schoukens is with the Control Systems group, Department of Electrical Engineering,
        Eindhoven University of Technology, 5612 AZ Eindhoven, The Netherlands
        {\tt\small m.schoukens@tue.nl}}%
}

\begin{document}

\maketitle
\thispagestyle{empty}
\pagestyle{empty}

\begin{abstract}
    The identification of black-box nonlinear state-space models requires a flexible representation of the state and output equation. Artificial neural networks have proven to provide such a representation. However, as in many identification problems, a nonlinear optimization problem needs to be solved to obtain the model parameters (layer weights and biases). A well-thought initialization of these model parameters can often avoid that the nonlinear optimization algorithm converges to a poorly performing local minimum of the considered cost function. This paper introduces an improved initialization approach for nonlinear state-space models represented as a recurrent artificial neural network and emphasizes the importance of including an explicit linear term in the model structure. Some of the neural network weights are initialized starting from a linear approximation of the nonlinear system, while others are initialized using random values or zeros. The effectiveness of the proposed initialization approach over previously proposed methods is illustrated on two benchmark examples.
\end{abstract}

\section{Introduction}
    Due to increasing performance demands, the introduction of more light-weight structures and increasing constraints on energy consumption of systems, nonlinear models and nonlinear control has become increasingly important \cite{Schoukens2019}. Since many nonlinearities, such as hysteresis or friction, are very difficult to model starting from first principles only, improved data-driven modelling approaches are required. A wide range of nonlinear modelling approaches and frameworks is available to the user: NARMAX, block-oriented and nonlinear state-space approaches to name a few \cite{Schoukens2019,Billings2013,SchoukensM2017b}.
    
    Due to their flexibility and their ability to model a wide range of nonlinear behavior, the identification of nonlinear state-space models has been of high interest over the last decade. This resulted in a variety of different approaches such as the polynomial nonlinear state-space method \cite{Paduart2010,Tiels2018}, fully probabilistic nonlinear state-space identification using particle filter approaches \cite{Schon2011,Schon2018}, subspace approaches \cite{Noel2013}, and state-space neural network (SS-NN) methods \cite{Suykens1995,Marconato2012,Forgione2019}.
    
    Artificial neural networks are known for their ability to represent multidimensional static nonlinear functions \cite{Barron1993}. This makes them a more suitable candidate than multivariate polynomial functions to represent the state and the output equation that are present in a nonlinear state-space model. Furthermore, \cite{Suykens1995} shows how a nonlinear state-space model can easily be represented as a recurrent neural network. 
    
    One of the main challenges in identifying a model which is nonlinear in the parameters, such as a SS-NN model, is the convergence to poorly performing local minima of the considered cost function. Both \cite{Suykens1995} and \cite{Paduart2010} propose an initialization approach based on a linear approximation of the nonlinear system. This paper further improves the initialization of SS-NN models starting from a linear approximation. Whereas \cite{Suykens1995} only discussed the idea of using an explicit linear part to increase the structure imposed on the model (e.g. the state transition function is nonlinearly dependent on the states and linearly dependent on the input), this paper demonstrates that the explicit inclusion of a linear layer in parallel to the nonlinear layers of the SS-NN improves the training behavior significantly. Furthermore the initialization of the layer weights and biases is revisited which also results in an improved training behavior.
    
    The remainder of the paper introduces the considered model class first (Section~\ref{sec:ModelClass}). Section~\ref{sec:SS-NN_Ident} introduces the proposed identification algorithm. The improved initialization approach is discussed in Section~\ref{sec:Init}. The improved performance of the proposed model class and the modified initialization approach is illustrated on two benchmark examples in Section~\ref{sec:Results}: the Bouc-Wen \cite{Noel2016,SchoukensM2017d} and the Wiener-Hammerstein \cite{Schoukens2009a} datasets.

\section{Model Class} \label{sec:ModelClass}
    \subsection{Nonlinear State-Space}
        The considered discrete-time nonlinear state-space model structure consists of a state and an output equation:
        \begin{align} \begin{split} \label{eq:ss}
            x(k+1) &= f(x(k),u(k)), \\
            y_0(k) &= g(x(k),u(k)),
        \end{split} \end{align}
        where $u(k) \in \mathbb{R}^{n_u \times 1}$, is the system input, and $y_0(k) \in \mathbb{R}^{n_y \times 1}$ is the noiseless system output, $x(k) \in \mathbb{R}^{n_x \times 1}$ denotes the system states and $k$ denotes the sample index. 
        
        One can separate a linear part in eq.~\eqref{eq:ss} without loss of generality:
        \begin{align} \begin{split} \label{eq:ss-L}
            x(k+1) &= A x(k) + B u(k) + \tilde{f}(x(k),u(k)), \\
            y_0(k) &= C x(k) + D u(k) + \tilde{g}(x(k),u(k)),
        \end{split} \end{align}
        with the matrices $A  \in \mathbb{R}^{n_x \times n_x}$, $B \in \mathbb{R}^{n_x \times n_u}$, $C \in \mathbb{R}^{n_y \times n_x}$, $D \in \mathbb{R}^{n_y \times n_u}$. The functions $\tilde{f}(\cdot)$ and $\tilde{g}(\cdot)$ are again static nonlinear functions. This split between nonlinear and linear dynamics automatically results when a polynomial representation is used for $f(\cdot)$ and $g(\cdot)$ \cite{Paduart2010}. Such a separation of the linear and nonlinear dynamics has also been reported in \cite{Forgione2019}, however, no results on the impact of explicitly including such a linear part have been reported there.
        
        In this work, the static nonlinearities are represented by a feedforward neural network with one hidden layer with $n_n$ neurons using a nonlinear activation function $\sigma(\cdot)$ and a linear output layer, e.g.:
        \begin{align} \label{eq:ANN-NL}
            f(x(k),u(k)) = W_x \sigma \left(\begin{bmatrix} W_{fx} W_{fu} \end{bmatrix} \begin{bmatrix} x(k) \\ u(k) \end{bmatrix} + b_f \right) + b_x,
        \end{align}
        where $\sigma(\cdot)$ is the nonlinear activation function which is often chosen as the hyperbolic tangent function or the radial basis function, $W_{fx}$, $W_{fu}$ and $W_x$ are the weights of the neural network,  $b_{fx}$, $b_{fu}$ and $b_x$ are the biases of the neural network, all with appropriate dimensions. A similar representation is used for $g(\cdot)$, $\tilde{f}(\cdot)$ and $\tilde{g}(\cdot)$.

	    A zero-mean, possibly colored, additive noise source $v(k)$ is assumed to be present at the output only $y(k) = y_0(k) + v(k)$. The noise source is assumed to have a finite variance. In case the noise is white, this corresponds to the classical output-error noise framework.
	
	\subsection{Uniqueness of the Parametrization}
	    The considered model representation is not unique. Beyond the well-known arbitrary state-space transformation that defines an equivalence class of models around \eqref{eq:ss} an \eqref{eq:ss-L}, also the neural network representation of the static nonlinearity is not uniquely parametrized. A simple permutation of the inner and other weights and biases can result in the same input-output behavior of the neural network \eqref{eq:ANN-NL}. Beyond the non-uniqueness of the state space and the static nonlinear functions separately, the linear terms denoted by the state-space matrices $A$, $B$, $C$ and $D$ can also be incorporated in the static nonlinear functions $\tilde{f}$ and $\tilde{g}$ in eq.~\eqref{eq:ss-L}.
    
    \subsection{State-Space Neural Network}
\begin{figure}[bt]
    		\centering
    			\includegraphics[width=0.9\columnwidth]{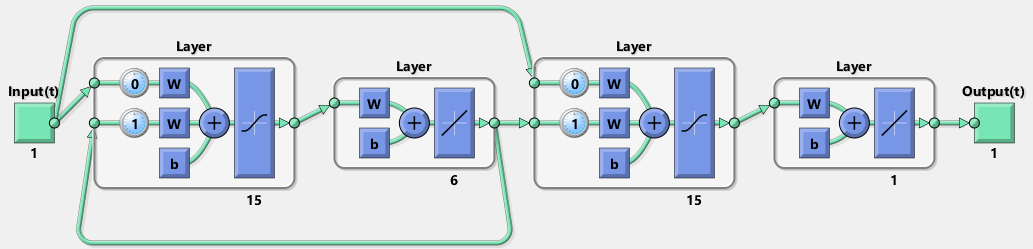}
    			\includegraphics[width=0.9\columnwidth]{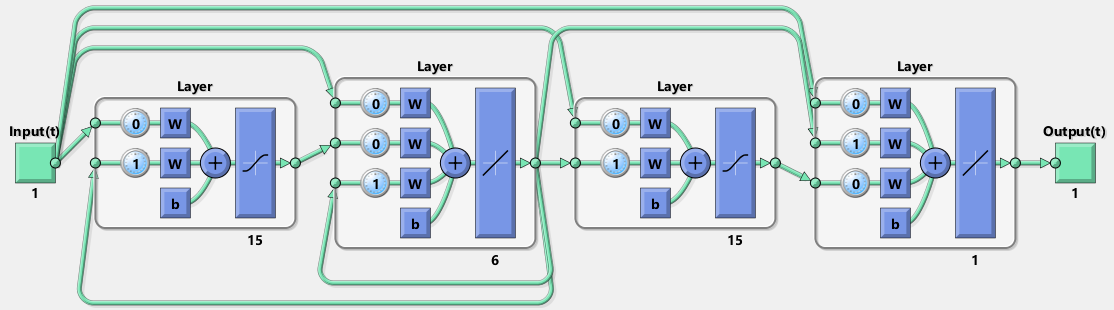}
    		\caption{A SS-NN representing the state-space equations of eq.~\eqref{eq:ss} (top) and eq.~\eqref{eq:ss-L} (bottom). The depicted SS-NN has 1 input, 1 output, 6 states and 15 neurons describing the state and output functions $f(\cdot)$ and $g(\cdot)$ (top) and $\tilde{f}(\cdot)$ and $\tilde{g}(\cdot)$ (bottom). The illustration is obtained using the deep learning toolbox in Matlab R2017b \cite{MatlabANN2017b}.}
    		\label{fig:SS-NN}
    	\end{figure}

        As is argued in \cite{Suykens1995}, a state-space representation using a neural network to represent the state and output equations can be interpreted as a specific form of a recurrent neural network, denoted as a state-space neural network (SS-NN). The recurrent neural networks representing eq.~\eqref{eq:ss} and eq.~\eqref{eq:ss-L} are depicted in Figure~\ref{fig:SS-NN}. Equation \eqref{eq:ss} becomes the following when written down as a recurrent neural network:
        \begin{align} \begin{split} \label{eq:SS-NN}
            x(k+1) &= W_x \sigma \left(\begin{bmatrix} W_{fx} & W_{fu} \end{bmatrix} \begin{bmatrix} x(k) \\ u(k) \end{bmatrix} + b_f \right) + b_x, \\
            y_0(k) &= W_y \sigma \left(\begin{bmatrix} W_{gx} & W_{gu} \end{bmatrix} \begin{bmatrix} x(k) \\ u(k) \end{bmatrix} + b_g \right) + b_y,
        \end{split} \end{align}
        the model type described by these equations are denoted as a state-space neural network (SS-NN). The same can be done for eq.~\eqref{eq:ss-L}, denoted as a generalized residual state-space neural network (gR-SS-NN):
        \begin{align} \begin{split} \label{eq:SS-NN-L}
            x(k+1) &= \begin{bmatrix} A & B \end{bmatrix}  \begin{bmatrix} x(k) \\ u(k) \end{bmatrix} \\ & \: + \tilde{W}_x \sigma \left(\begin{bmatrix} \tilde{W}_{fx} & \tilde{W}_{fu} \end{bmatrix} \begin{bmatrix} x(k) \\ u(k) \end{bmatrix} + \tilde{b}_{f} \right) + \tilde{b}_x, \\
            y_0(k) &= \begin{bmatrix} C & D \end{bmatrix}  \begin{bmatrix} x(k) \\ u(k) \end{bmatrix} \\ & \: + \tilde{W}_y \sigma \left(\begin{bmatrix} \tilde{W}_{gx} & \tilde{W}_{gu} \end{bmatrix} \begin{bmatrix} x(k) \\ u(k) \end{bmatrix} + \tilde{b}_{g} \right) + \tilde{b}_y,
        \end{split} \end{align}

        Observe that the inclusion of an explicit linear part in eq.~\eqref{eq:ss-L} and \eqref{eq:SS-NN-L} can be interpreted as a generalized form of a so-called residual network or resNet \cite{He2016}. Instead of including a unity weighting matrix in the linear layers, a fully connected linear weighting is employed in this paper to model the 'linear' dynamics, the residuals that are present on top of these linear dynamics are captured by the neural network layers that have nonlinear activation functions (layers 1 and 3 in the bottom Figure~\ref{fig:SS-NN}). It is illustrated in this paper that, together with an improved initialization, the inclusion of the explicit linear part improves the estimation of the SS-NN model significantly.
        
\section{Estimating a SS-NN Model} \label{sec:SS-NN_Ident}
    \subsection{Cost Function Optimization}
        The model parameters are obtained as the minimization of the mean squared simulation error:
        \begin{align} \label{eq:cost}
            V_N(\theta) = \frac{1}{N} \sum_{k=1}^N \left( y(k) - \hat{y}(k|\theta) \right)^2,
        \end{align}
        \begin{align} \label{eq:theta}
            \hat{\theta} = \underset{\theta}{\argmin} V_N(\theta),
        \end{align}
        where $\hat{y}(k|\theta)$ is the simulated output of the SS-NN model given the parameter vector $\theta$ and $y(k)$ is the noisy system output. The parameter vector $\theta$ contains all the model parameters: the state-space matrix entries and the weights and biases of the neural network representing the static nonlinearities $f(\cdot)$ and $g(\cdot)$ or $\tilde{f}(\cdot)$ and $\tilde{g}(\cdot)$ respectively. $N$ represents the total number of samples over which the cost function is computed.
        
        Eq.~\eqref{eq:cost} is nonlinear in the parameters and typicality not convex but its gradients can be efficiently computed. Hence the Levenberg-Marquardt algorithm \cite{Levenberg1944}, a gradient descent-type algorithm, is used to minimize the cost function. The required gradients can efficiently be obtained using the backpropagation-through-time algorithm \cite{Werbos1988}. A 'data-driven coordinate frame' is used to get rid off equivalent gradient directions (resulting in a rank deficient Jacobian) due to the non-uniqueness of the model representation. In practice this is achieved using the singular value decomposition of the Jacobian \cite{Wills2008}.
        
\section{Initialization of SS-NN} \label{sec:Init}
    Due to the use of the Levenberg-Marquardt (or any other gradient-based) optimization algorithm, an initial guess of the parameter values is required. The subsections below describe the use of 3 different parameter initialization schemes.
    
    The initialization is performed under the assumption that the input signal is zero mean and has been normalized to have a standard deviation equal to 1.
    
    \subsection{Random Initialization}
        Random initialization of the parameters is the standard approach in most neural network training algorithms. However, such a random initialization needs to be performed with caution as one can easily obtain unstable initial models, which most often lead to bad final estimates, or simply cause the algorithm to stop the optimization due to exploding gradients, or due to an exploding model output.
        
        Here we consider the random initialization of the layer weights and biases using a uniform distribution with a support from -1 to 1 (denoted by $\mathcal{U}(-1,1)$) or zeros. Except for some layer weights of eq.~\eqref{eq:SS-NN-L}, they are initialized as zeros to avoid unstable initial estimates. The uniform random initialization of the parameters is the common approach when training a neural network \cite{Bishop1995}. Table~\ref{table:randomInit} gives a detailed overview of the random weights and bias initialization scheme. The normalization of some weights with $\sqrt{n_x}$ or $\sqrt{n_u}$ ensures that the input of the nonlinear activation function does not grow too large for an increasing state or input dimension.
        
            \begin{table}[ht]
                \centering
                \caption{Random weights and biases initialization. All weights and bias matrices are of appropriate dimension.}
                \begin{tabular}{c | c c} 
                 \rule{0pt}{2ex} 
                           & SS-NN  & gR-SS-NN \\
                 \hline 
                 $W_x$  & $\mathcal{U}(-1,1)$ & $\mathcal{U}(-1,1)$ \\
                 $W_{fx}$   & $\mathcal{U}(-1,1) / \sqrt{n_x}$ & $\mathcal{U}(-1,1) / \sqrt{n_x}$\\
                 $W_{fu}$ & $\mathcal{U}(-1,1)/ \sqrt{n_u}$ & $\mathcal{U}(-1,1)/ \sqrt{n_u}$\\
                 $W_y$ & $\mathcal{U}(-1,1)$ & $0$ \\
                 $W_{gx}$ & $\mathcal{U}(-1,1) / \sqrt{n_x}$ & $\mathcal{U}(-1,1) / \sqrt{n_x}$ \\
                 $W_{gx}$ & $\mathcal{U}(-1,1)/ \sqrt{n_u}$ & $\mathcal{U}(-1,1)/ \sqrt{n_u}$ \\
                 $b_x$ & 0 & 0 \\
                 $b_{f}$ & 0 & 0 \\
                 $b_y$ & 0 & 0 \\
                 $b_{g}$ & 0 & 0 \\
                 $A$  & / & 0 \\
                 $B$  & / & $\mathcal{U}(-1,1)$ \\
                 $C$  & / & $\mathcal{U}(-1,1)$ \\
                 $D$  & / & $\mathcal{U}(-1,1)$ \\
                \end{tabular}
                 \label{table:randomInit}
            \end{table}
        
    \subsection{Initialization using a Linear Approximation}
        An alternative initialization scheme was proposed in \cite{Suykens1995} for a SS-NN following eq.~\eqref{eq:ss}. This scheme makes use of the well known LTI identification tools that are available in the literature \cite{Ljung1999,Pintelon2012}. These tools can obtain a linear approximation of the nonlinear system, expressed in the state-space matrices $A_{LTI}$, $B_{LTI}$, $C_{LTI}$, $D_{LTI}$. The Best Linear Approximation framework \cite{Pintelon2012,Schoukens2016,Enqvist2005} is a well established framework that can guide the practitioner in obtaining such a linear approximate model of a nonlinear system. The state-space matrices of the linear approximate model are normalized such that each of the states has a standard deviation equal to 1 \cite{SchoukensM2020}.
        
        \subsubsection{Random Weights in the Linear Layers}
        Suykens and co-authors \cite{Suykens1995} observed that such a linear initial guess can be incorporated in a SS-NN by making use of the linear regime of some typical activation functions such as the hyperbolic tangent function if the number of neurons in the nonlinear layers is larger than the number of states in the system-to-be-modeled. Table~\ref{table:LtiInit} details the initialization of a SS-NN using a previously obtained LTI approximation of the nonlinear system, represented in state-space form. The factor $\gamma$ is chosen such that the nonlinear activation function operates in a linear regime for the given training dataset. Additional random values are used to initialize some of the weights of the layers with linear activation functions, while still ensuring that the initial model behaves as the linear approximation we started from.
        
        \subsubsection{Random Weights in the Nonlinear Layers}
        Observe that \cite{Suykens1995} initializes the linear layer with random numbers, and uses zero weights for the nonlinear layer weights that are not initialized by the linear approximation. This often results in a suboptimal initial estimate. This paper proposes to do this the other way around, random weights in the nonlinear layer and zero weights in the linear layers, works better in the benchmark examples in Section~\ref{sec:Results} of this paper. Intuitively one can understand this by observing that using random weights and biases in the nonlinear layer generates a pool of nonlinearly transformed outputs which the estimator can pick from using the linear weights during optimization. Again, the initialization scheme is detailed in Table~\ref{table:LtiInit}.
        
        \subsubsection{Explicit Linear Dynamics - generalized Residual Network}
        The gR-SS-NN structure offers a natural way to initialize the neural network using a linear approximation. The linear state-space matrices are directly used to initialize the $A$, $B$, $C$, $D$ matrices in eq.~\eqref{eq:SS-NN-L}. This leaves quite some flexibility in initializing the weights and biases of the nonlinear layers. As in the previous initialization approach, random weights and biases in the nonlinear layer generates a pool of nonlinearly transformed outputs which the estimator can pick from using the linear weights during optimization. The the full initialization scheme is detailed in Table~\ref{table:LtiInit}.
        
        Since this initialization scheme does not make explicit use of the linear region of the chosen activation function, a broader set of activation functions, such as radial basis functions or rectified linear units, can be considered for the gR-SS-NN structure.
        
\begin{table}[ht]
    \centering
    \caption{Initialization starting from a linear approximation. All weights and bias matrices are of appropriate dimension.}
    \begin{tabular}{c | c c c} 
     \rule{0pt}{2ex} 
               & SS-NN & SS-NN & gR-SS-NN \\
               & following \cite{Suykens1995}  & this paper & this paper \\
     \hline 
     $W_x$  & $\frac{1}{\gamma}[I_{n_x} \: \: \mathcal{U}(-1,1)]$ & $\frac{1}{\gamma}[I_{n_x} \: \: 0]$ & $0$ \\
     $W_{fx}$   & $\begin{bmatrix}\gamma A_{LTI} \\ 0 \end{bmatrix}$ & $\begin{bmatrix}\gamma A_{LTI} \\ \mathcal{U}(-1,1)/\sqrt{n_x} \end{bmatrix}$ & $\mathcal{U}(-1,1) / \sqrt{n_x}$ \\
     $W_{fu}$    & $\begin{bmatrix}\gamma B_{LTI} \\ 0 \end{bmatrix}$ & $\begin{bmatrix}\gamma B_{LTI} \\ \mathcal{U}(-1,1) / \sqrt{n_u} \end{bmatrix}$ & $\mathcal{U}(-1,1)/ \sqrt{n_u}$ \\
     $W_y$ & $\frac{1}{\gamma}[I_{n_y} \quad \mathcal{U}(-1,1)]$ & $\frac{1}{\gamma}[I_{n_y} \quad 0]$ & $0$ \\
     $W_{gx}$ & $\begin{bmatrix}\gamma C_{LTI} \\ 0 \end{bmatrix}$ & $\begin{bmatrix}\gamma C_{LTI} \\ \mathcal{U}(-1,1)/\sqrt{n_x} \end{bmatrix}$ & $\mathcal{U}(-1,1) / \sqrt{n_x}$ \\
     $W_{gx}$  & $\begin{bmatrix}\gamma D_{LTI} \\ 0 \end{bmatrix}$ & $\begin{bmatrix}\gamma D_{LTI} \\ \mathcal{U}(-1,1) / \sqrt{n_u} \end{bmatrix}$ & $\mathcal{U}(-1,1)/ \sqrt{n_u}$\\
     $b_x$  & 0 & 0 & 0 \\
     $b_{f}$  & 0 & $\begin{bmatrix} 0_{n_x} \\ \mathcal{U}(-1,1) \end{bmatrix}$ & $\mathcal{U}(-1,1)$ \\
     $b_y$  & 0 & 0 & 0\\
     $b_{g}$  & 0 & $\begin{bmatrix} 0_{n_x} \\ \mathcal{U}(-1,1) \end{bmatrix}$ & $\mathcal{U}(-1,1)$\\
     $A$  & / & / & $A_{LTI}$\\
     $B$  & / & / & $B_{LTI}$\\
     $C$  & / & / & $C_{LTI}$\\
     $D$  & / & / & $D_{LTI}$\\
    \end{tabular}
     \label{table:LtiInit}
\end{table}
    
\section{Benchmark Results} \label{sec:Results}
    The proposed model structures are compared on two different nonlinear system benchmarks: the Bouc-Wen \cite{Noel2016,SchoukensM2017d} and the Wiener-Hammerstein \cite{Schoukens2009a} datasets. The effectiveness of the different initialization schemes are studied using a Monte-Carlo simulation on both benchmark datasets. The models are trained using the Matlab Deep Learning toolbox. The developed algorithm is available through: \url{https://gitlab.tue.nl/mschouke/ss-nn}.
    
    \subsection{Bouc-Wen Benchmark}
        \subsubsection{System and Data} 
            The Bouc-Wen system is a mass-spring-damper system with a hysteretic nonlinearity. It is available as a simulation script allowing users to generate their own data for identification. We used one period of a random phase multisine input signal \cite{Pintelon2012}. The signal was $N=8192$ samples long and it excites the full frequency grid between 5 Hz and 150 Hz. The input signal amplitude is 50 N\textsubscript{rms}.
            
            Two test datasets are available for the Bouc-Wen benchmark system: a multisine and a sinesweep dataset. The multisine test output is obtained as the steady-state response of the system excited by a random phase multisine of $N=8192$ samples long, with the same frequency range and amplitude as the input used during estimation. The sinesweep test output is obtained by exciting the system, starting from zero initial conditions, with a sinesweep signal with an amplitude of 40  N\textsubscript{rms}. The frequency band from 20 to 50 Hz is covered at a sweep rate of 10 Hz/min.
            
            A detailed description of the Bouc-Wen benchmark system is provided in \cite{Noel2016,SchoukensM2017d}, the data can be obtained from \url{www.nonlinearbenchmark.org}.
        
        \subsubsection{Model Settings}
            A 3rd order state-space model ($n_x = 3$) is used, in line with previous findings \cite{Noel2017b}. A total of 15 neurons ($n_n = 15$, tansig activation function) are used to represent the static nonlinearities of both the state and the output equation.
            
            The LTI approximation used to initialize the weights and biases of the SS-NN is obtained by the time-domain prediction error method as implemented in Matlab by the function {\tt ssest}. This function initializes the parameter estimates using either a subspace approach or an iterative rational function estimation approach. The state-space matrices are refined subsequently using the prediction error minimization approach \cite{Ljung1999}.
        
        \subsubsection{Results}
            As a measure of model quality, the simulation RMSE (root mean squared error) on the multisine and sinesweep test dataset is reported:
            \begin{align} \label{eq:RMSE}
                e_{\text{RMSE}} = \sqrt{\frac{1}{N} \sum_{k=1}^N \left( y(k) - \hat{y}(k|\theta) \right)^2},
            \end{align}
            where $y(k)$ is the observed test output and $\hat{y}(k|\theta)$ is the simulated output using the estimated model. To make sure the model output is in steady state for the multisine dataset two periods are simulated and the RMSE is calculated on the second period. The sinesweep data is not in steady state, hence, the first 2000 samples are ignored when calculating the RMSE to allow the transient to decay.
            
            Since all of the proposed initialization schemes contain some random initial parameters, a Monte Carlo simulation has been performed of 100 Monte Carlo runs.
            
\begin{figure}[bt] 
        		\centering
        		    \includegraphics[width=0.90\columnwidth]{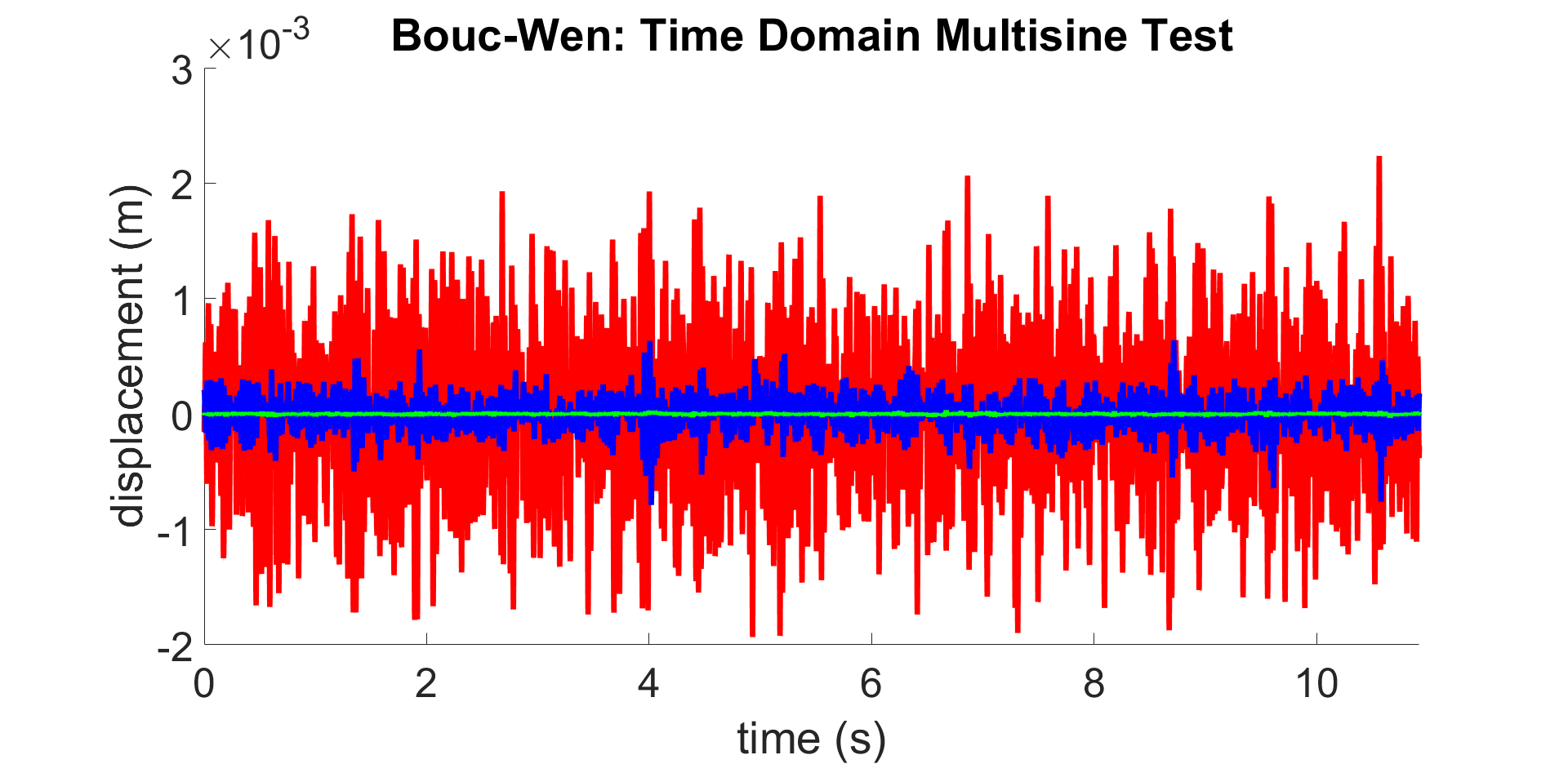}
        			\includegraphics[width=0.90\columnwidth]{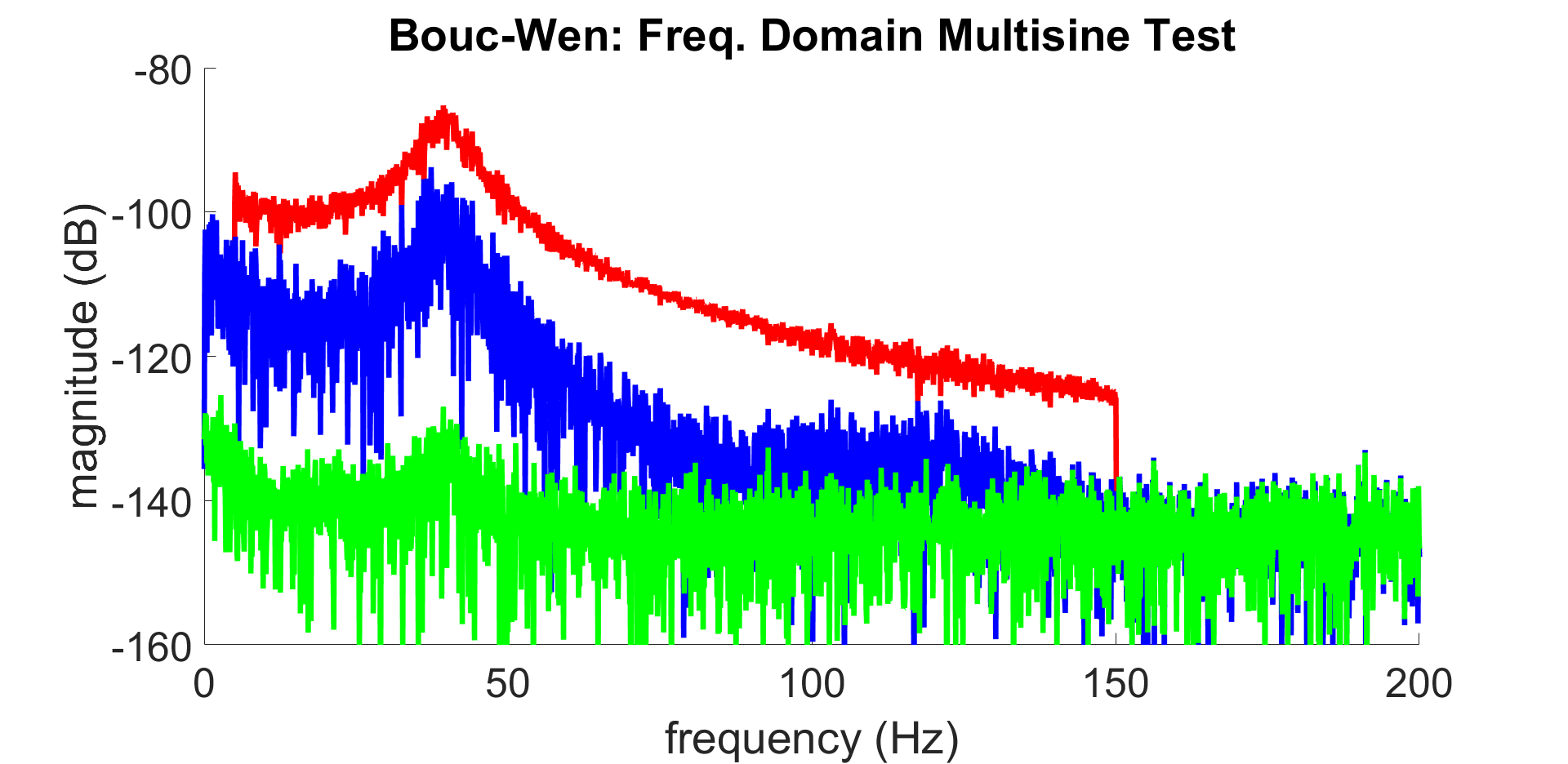}
        		\caption{Time- and frequency-domain test results of the median gR-SS-NN model initialized using a linear approximation on the Bouc-Wen multisine test data. The true system output is shown in red, the residuals obtained with a 3rd order LTI model are shown in blue and the residuals obtained with the gR-SS-NN model are shown in green.}
        		\label{fig:BW-MS}
        	\end{figure}

            The obtained gR-SS-NN models using linear approximation initialization obtain state of the art results, with an error which is almost half of the one reported in \cite{Noel2017b} ($7.14 \; 10^{-6}$ versus $1.2 \; 10^{-5}$ RMSE on the multisine test data). The error in both the time- and frequency domain of the median gR-SS-NN model (in RMSE sense) is shown in Figure~\ref{fig:BW-MS} for the multisine test signal. These figures also illustrates that the obtained has an RMSE of almost a factor 100 lower than a LTI model.
            
\begin{figure}[bt] 
        		\centering
        		    \includegraphics[width=0.9\columnwidth]{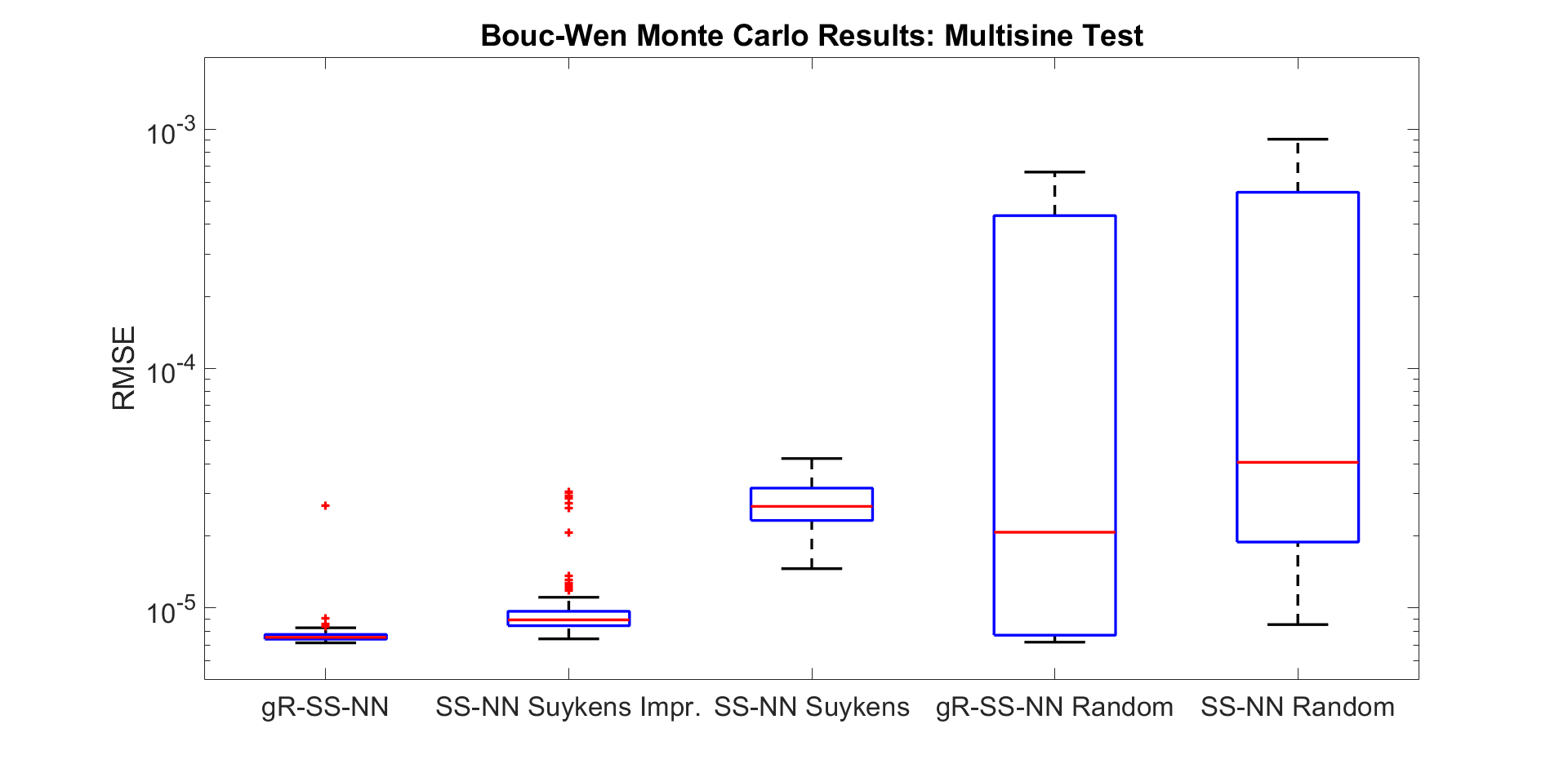}
        			\includegraphics[width=0.9\columnwidth]{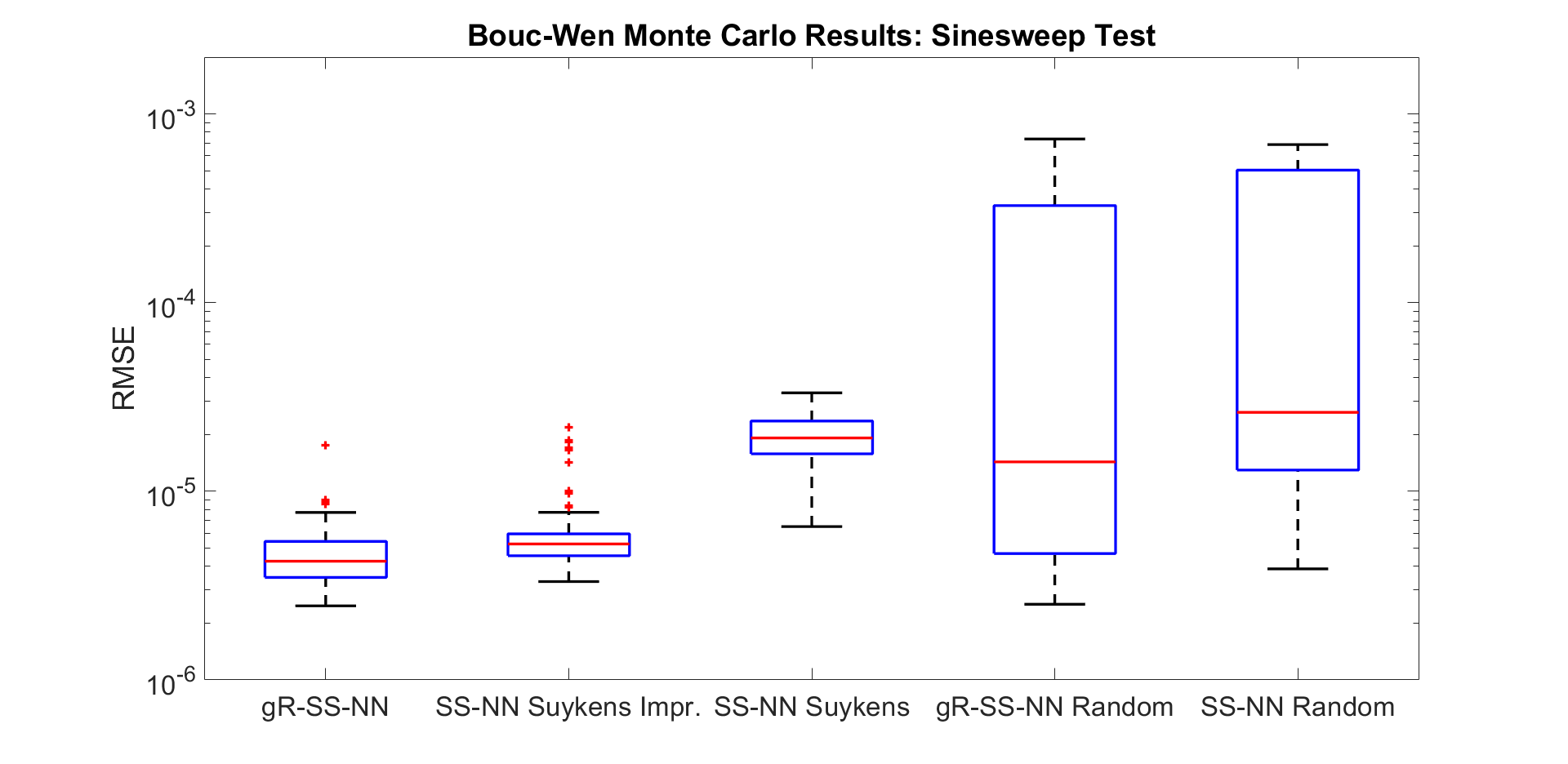}
        		\caption{Boxplots on a logarithmic scale of the test results for both the multisine (top) and the sinesweep (bottom) test data of the Monte Carlo obtained SS-NN models using the various initialization methods: gR-SS-NN LTI init. (Table~\ref{table:LtiInit}, column 3), SS-NN LTI improved init. (Table~\ref{table:LtiInit}, column 2), gR-SS-NN LTI init. \cite{Suykens1995} (Table~\ref{table:LtiInit}, column 1), gR-SS-NN Random init. (Table~\ref{table:randomInit}, column 2), SS-NN Random init. \cite{Suykens1995} (Table~\ref{table:randomInit}, column 1).}
        		\label{fig:BW-Boxplot}
        	\end{figure}

            Furthermore, the gR-SS-NN model with linear approximation initialization outperforms all the other SS-NN models using various initialization schemes. This can be observed from the boxplots shown in Figure~\ref{fig:BW-Boxplot}. The fully random initialization methods occasionally result in a high quality estimate, but the median and the spread of the resulting models is very large. The three models based on a linear approximate model have a much smaller spread, and in the case of the gR-SS-NN and the SS-NN improved initialization (Table~\ref{table:LtiInit}, column 3 and 2 respectively) this results also in a much lower median RMSE. 
            
\begin{figure}[bt] 
        		\centering
        			\includegraphics[width=0.9\columnwidth]{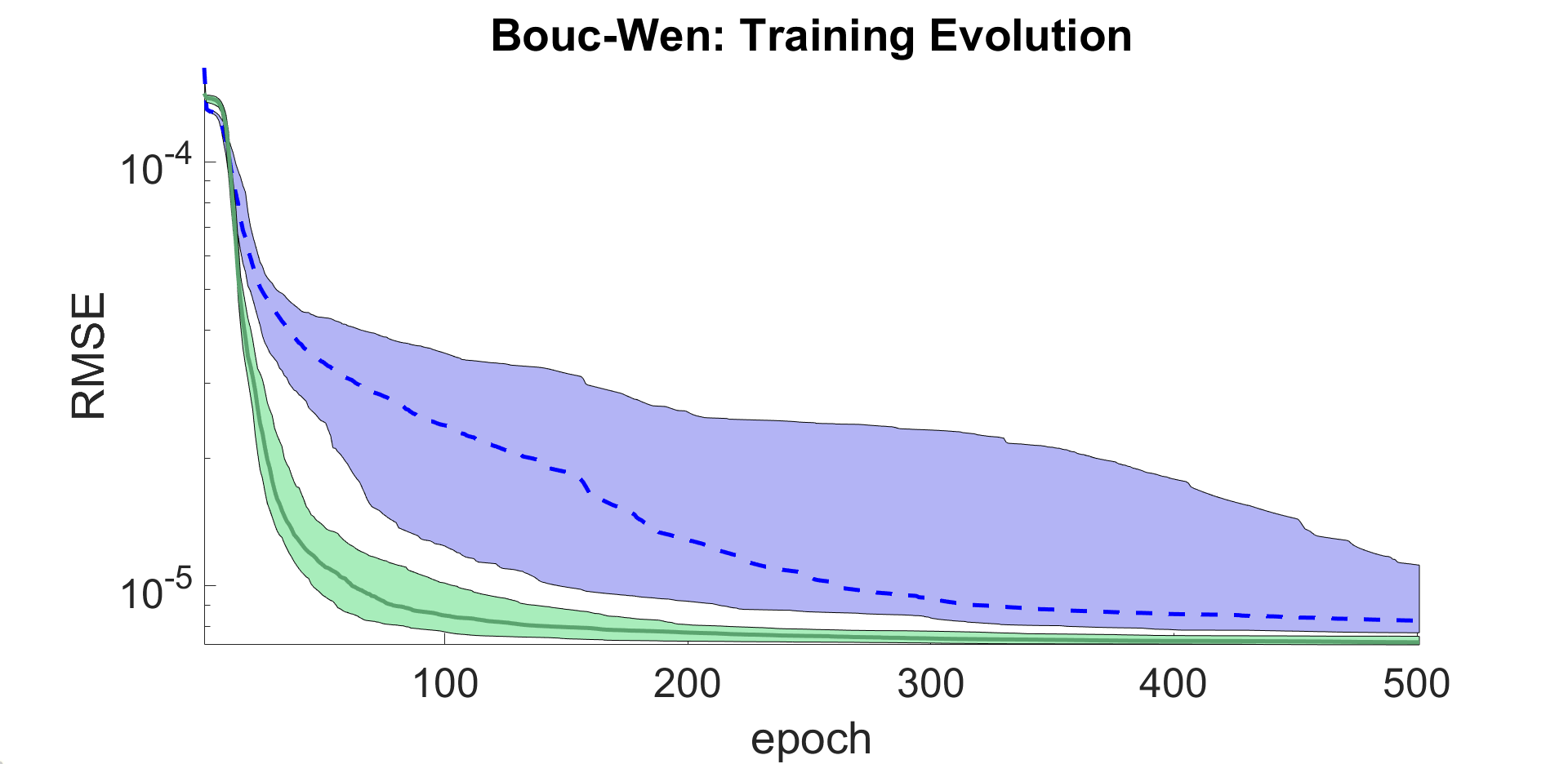}
        			\includegraphics[width=0.9\columnwidth]{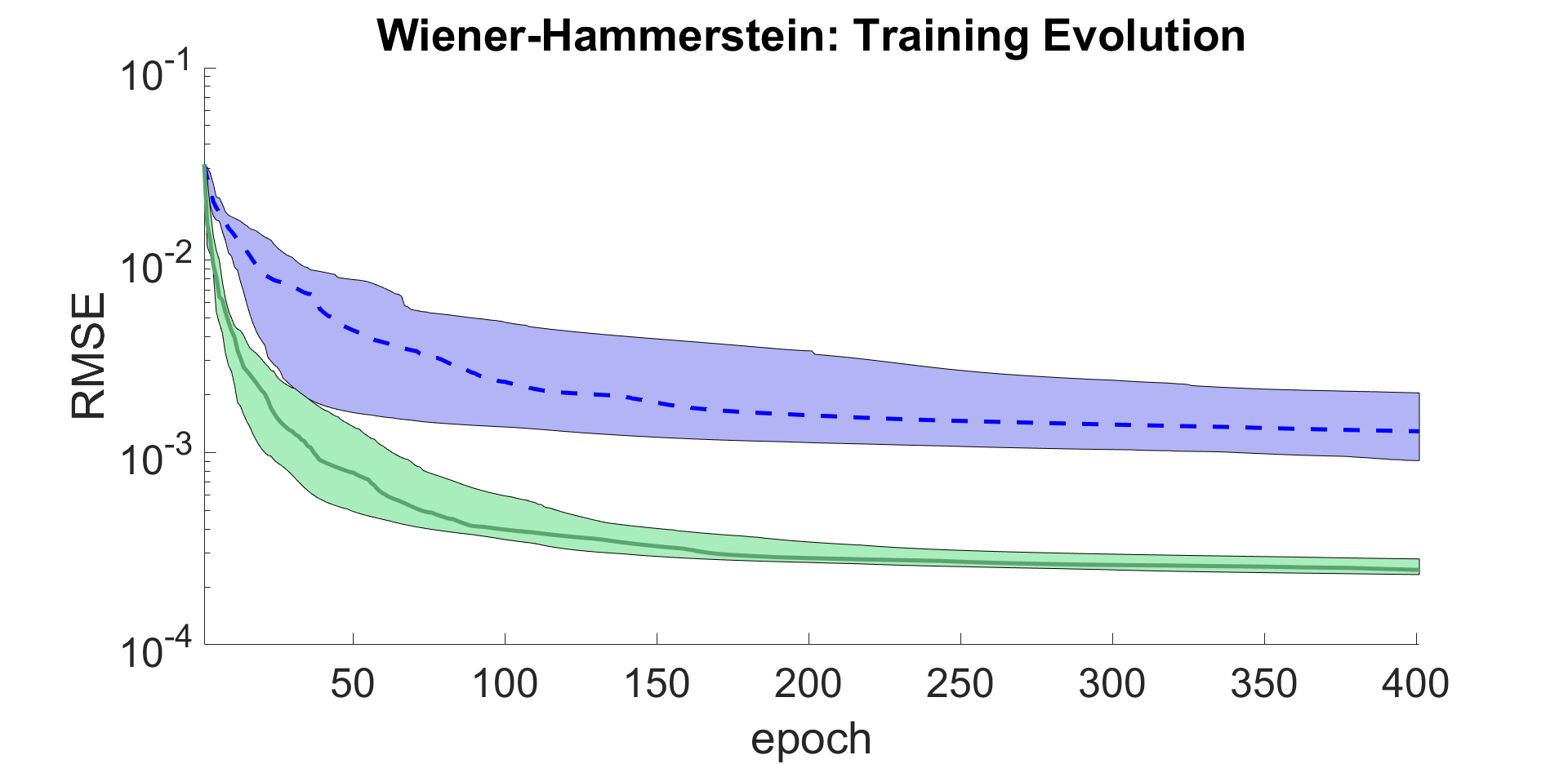}
        		\caption{Evolution of the RMSE on the training dataset over the training epochs for both the Bouc-Wen and the Wiener-Hammerstein benchmark. The thick green line visualizes the median error of the training of a gR-SS-NN model with a linear approximation initialization (Table~\ref{table:LtiInit}, column 3), the green zone around it shows the 10\% - 90\% region of the RMSE. The thick blue dashed line and the colored zone around visualizes the same for the training of a SS-NN model using the improved linear approximation initialization (Table~\ref{table:LtiInit}, column 2).}
        		\label{fig:BW-Training}
        		\vspace{-0.5 cm}
        	\end{figure}

            The introduction of the generalized residual network, in other words the explicit inclusion of linear dynamics in the SS-NN, clearly improves the model behavior for both the random and linear approximate-based initialization methods. This can both be observed from the boxplots in  Figure~\ref{fig:BW-Boxplot}, but also in the evolution of the RMSE over the training epochs shown in Figure~\ref{fig:BW-Training}. The RMSE of the gR-SS-NN models starting from a linear approximation decay much faster, resulting in the need for much less training epochs. Since the gR-SS-NN models only have slightly higher complexity compared to the SS-NN models, the training time can be reduced significantly.
            
    \subsection{Wiener-Hammerstein Benchmark}
        \subsubsection{System and Data} 
            The Wiener-Hammerstein benchmark \cite{Schoukens2009a} features a Wiener-Hammerstein block-oriented electronic circuit: a one-sided saturation diode-resistor nonlinearity is sandwiched in between two 3rd-order low-pass linear filters. Only noise at the output of the system is present (SNR$\approx$60dB). The training and test input data consist of a filtered white Gaussian noise sequence (10kHz cut-off frequency). The benchmark data consist of  and more information on the benchmark system and previous benchmark results is available through: \url{www.nonlinearbenchmark.org}.
        
        \subsubsection{Model Settings}
            A 6th order state-space model ($n_x = 6$) is used, in line with the system description \cite{Schoukens2009a}. A total of 15 neurons ($n_n = 15$, tansig activation function) are used to represent the static nonlinearities of both the state and the output equation.
            
            The LTI approximation used to initialize the weights and biases of the SS-NN is obtained in the same way as outlined for the Bouc-Wen benchmark results.
        
        \subsubsection{Results}
\begin{figure}[bt] 
        		\centering
        		    \includegraphics[width=0.9\columnwidth]{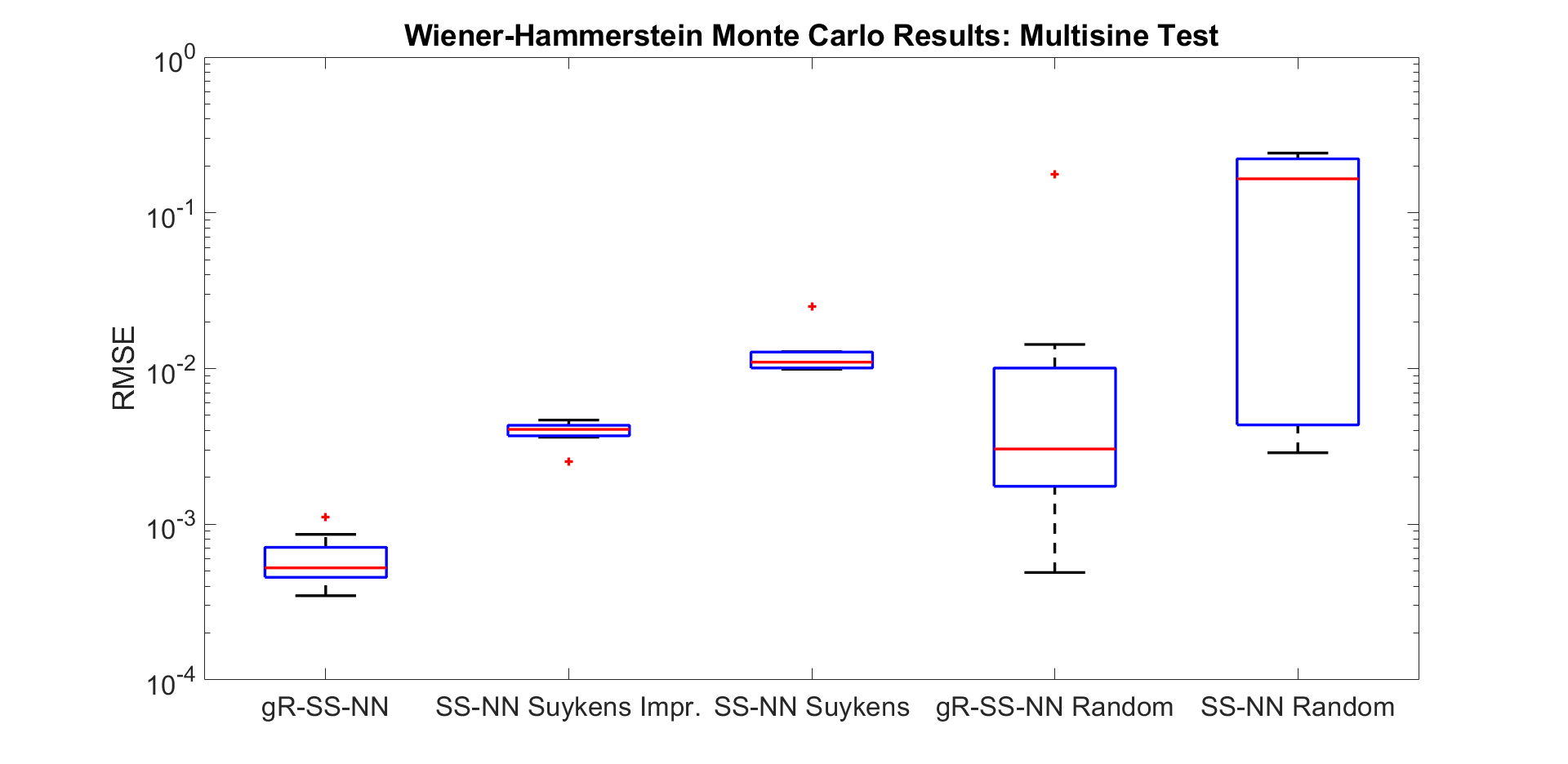}
        		\caption{Boxplots of the test results of the Monte Carlo simulation obtained SS-NN models using the various initialization methods: gR-SS-NN LTI init. (Table~\ref{table:LtiInit}, column 3), SS-NN LTI improved init. (Table~\ref{table:LtiInit}, column 2), gR-SS-NN LTI init. \cite{Suykens1995} (Table~\ref{table:LtiInit}, column 1), gR-SS-NN Random init. (Table~\ref{table:randomInit}, column 2), SS-NN Random init. \cite{Suykens1995} (Table~\ref{table:randomInit}, column 1).}
        		\label{fig:WH-Boxplot}
        		\vspace{-0.5 cm}
        	\end{figure}

            The general observations are in line with the observations obtained for the Bouc-Wen benchmark: the use of an linear approximation to initialize the SS-NN model greatly reduces the spread of the quality of the obtained models, while the introduction of the explicit linear part (generalized residual net) lowers the RMSE obtained by these models significantly (Figure~\ref{fig:WH-Boxplot}) and reduces the training time (Figure~\ref{fig:BW-Training}). 
            
\begin{figure}[bt] 
        		\centering
        		    \includegraphics[width=0.90\columnwidth]{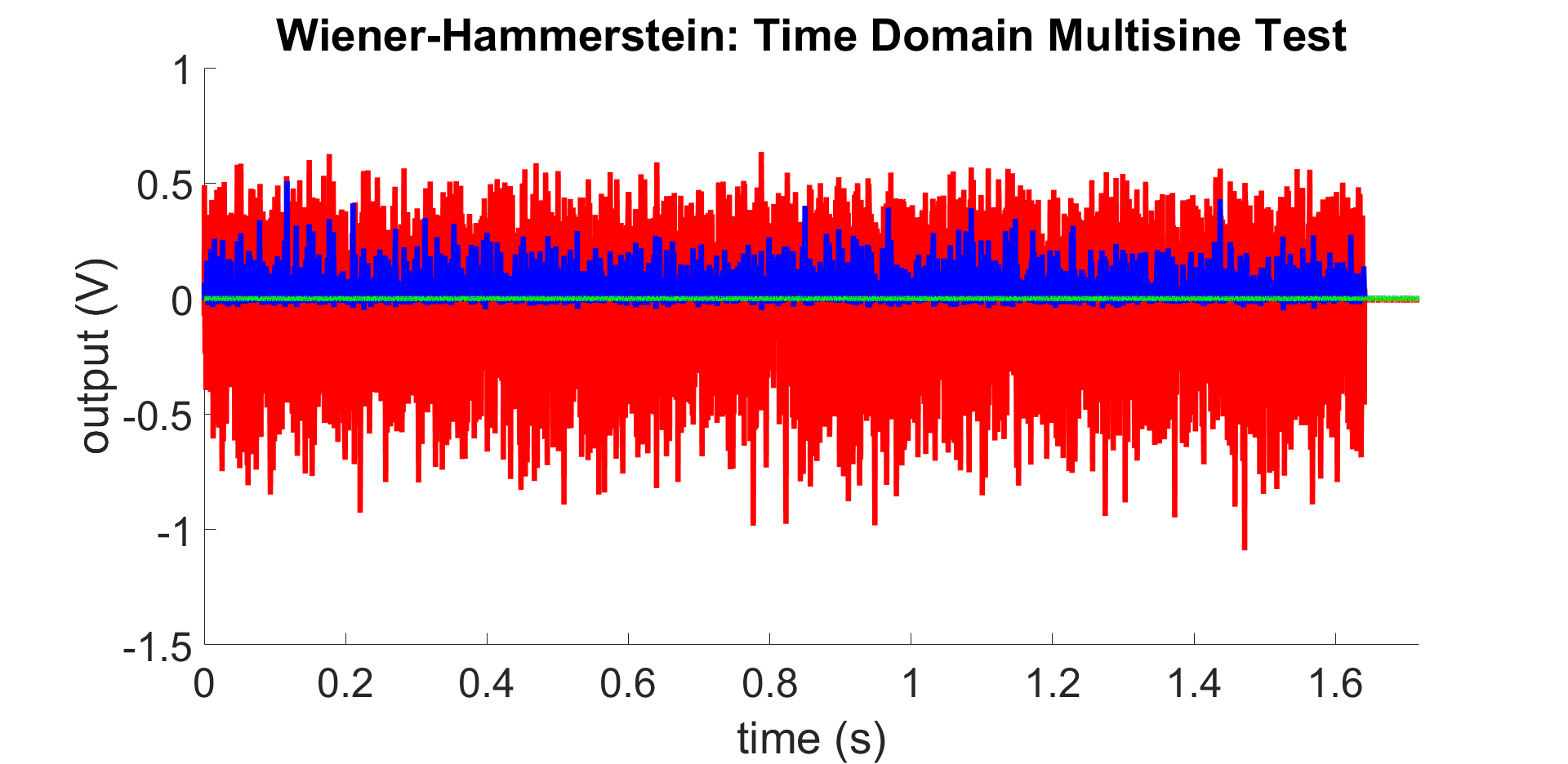}
        			\includegraphics[width=0.90\columnwidth]{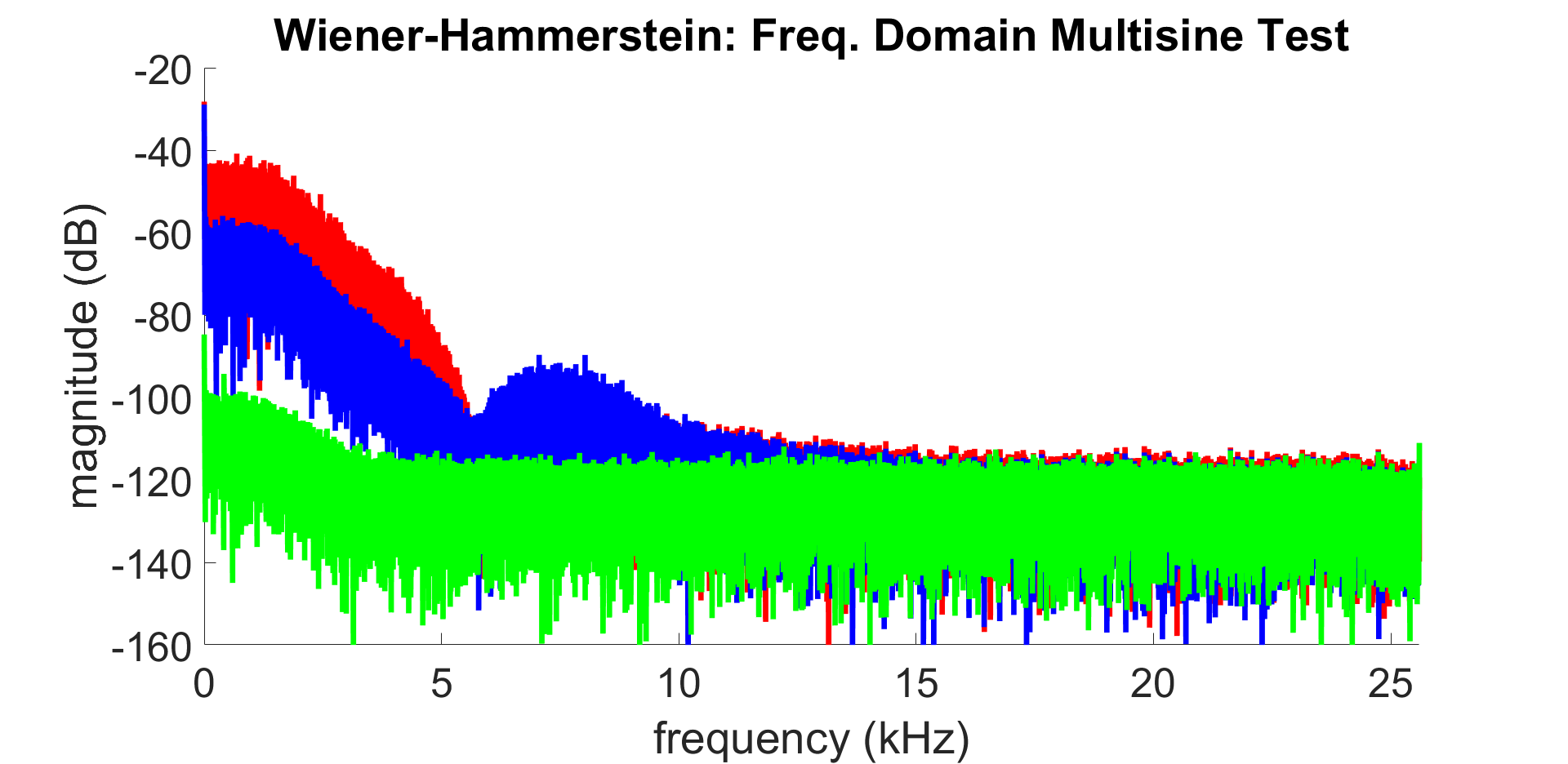}
        		\caption{Time- and frequency-domain test results of the median gR-SS-NN model initialized using a linear approximation on the Wiener-Hammerstein test data. The true system output is shown in red, the residuals obtained with a 6th order LTI model and with the gR-SS-NN model are shown in bule and green respectively.}
        		\label{fig:WH-MS}
        	\end{figure}

            Furthermore, it can be observed that the best obtained RMSE (0.34 mV) is quite close to the best result reported in the literature (0.278 mV, \cite{SchoukensM2014}). The performance of a gR-SS-NN model with the median RMSE can be observed in Figure~\ref{fig:WH-MS}. This figures also illustrates that the best obtained model has an RMSE of almost a factor 100 lower than a LTI model.
        
\section{Conclusions}
    This paper proposed two important improvements in the initialization of state-space neural network models for dynamical systems, compared to the existing approaches outlined in the literature. Firstly, starting from a linear approximate model, some weights and biases of the SS-NN can be initialized, while the other weights are initialized by random numbers. It turns out to be valuable to randomly initialize the weights and biases of the nonlinear layers, rather than the linear ones. Secondly, the inclusion of an explicit linear term in parallel to the nonlinear function results in a significantly improved training behavior, the resulting SS-NN is denoted as a generalized residual state-space neural network. This is in line with the positive results obtained using residual neural networks in the machine learning literature. These observations have been obtained by studying Monte-Carlo simulation results on two well-established benchmark systems.

\bibliographystyle{unsrt}
\bibliography{references}

\end{document}